\documentclass[11pt,a4paper]{article}
\usepackage{times,latexsym}
\usepackage{url}
\usepackage[T1]{fontenc}
\usepackage{enumitem}
\usepackage{times}
\usepackage{url}
\usepackage{latexsym}
\usepackage{bm}
\usepackage{booktabs}
\usepackage{amsmath}
\usepackage{paralist}
\usepackage{verbatim}
\usepackage{graphicx} 
\usepackage{multirow}
\usepackage{verbatim}

\usepackage[acceptedWithA]{tacl2018v2}
\usepackage{times}
\usepackage{url}
\usepackage{latexsym}
\usepackage{amsmath}
\usepackage{bm}

\usepackage{xspace,mfirstuc,tabulary}

\newcommand{\semfit}{\textsc{Semfit}}

\newif\iftaclinstructions
\taclinstructionsfalse 
\iftaclinstructions
\renewcommand{\confidential}{}
\renewcommand{\anonsubtext}{(No author info supplied here, for consistency with
TACL-submission anonymization requirements)}
\newcommand{\instr}
\fi

\iftaclpubformat 

\else

\fi

\title{Efficient Contextual Representation Learning Without Softmax Layer}

\author{
 Liunian Harold Li$^\dagger$, Patrick H. Chen$^*$, Cho-Jui Hsieh$^*$, Kai-Wei Chang$^*$\\
 $^\dagger$Peking University \\
 $^*$University of California, Los Angeles\\
 \texttt{liliunian@pku.edu.cn, patrickchen@g.ucla.edu}\\
 \texttt{\{chohsieh, kwchang\}@cs.ucla.edu} \\
}
\date{}

\begin{document}
\maketitle

\begin{abstract}
  Contextual representation models have achieved great success in improving various downstream tasks. However, these language-model-based encoders are difficult to train due to the large parameter sizes and high computational complexity. By carefully examining the training procedure, we find that the softmax layer (the output layer) causes significant inefficiency due to the large vocabulary size. Therefore, we redesign the learning objective and propose an efficient framework for training contextual representation models. Specifically, the proposed approach bypasses the softmax layer by performing language modeling with dimension reduction, and allows the models to leverage pre-trained word embeddings.
  Our framework reduces the time spent on the output layer to a negligible level, eliminates almost all the trainable parameters of the softmax layer and performs language modeling without truncating the vocabulary.
  When applied to ELMo, our method achieves a 4 times speedup and eliminates 80\% trainable parameters while achieving competitive performance on downstream tasks.
\end{abstract}

\section{Introduction}
In recent years, text representation learning approaches, such as ELMo \cite{elmo}, GPT-1 \cite{GPT}, BERT \cite{bert} and GPT-2 \cite{GPT2}, have been developed to provide generic contextual representations for the natural language, which have led to large improvements to various downstream tasks. The key idea underneath is to train a contextual encoder with a language model objective on a large unannotated text corpus.
During the training, part of the text is masked and the goal is to encode the remaining context and predict the missing part. Specifically, given a corpus with vocabulary size $V$, the model consists of two parts: 1) an encoder is learned to embed the context and output an $m$-dimensional vector, and 2) the output vector goes through a softmax layer, where it is multiplied by an embedding matrix $\bm{W} \in \mathcal{R}^{V\times m}$ and is fed into a softmax function, to produce a conditional distribution of the missing word. The encoders trained in such a way are able to capture generic contextual information of the input text and have been used in a variety of benchmark tasks to establish state-of-the-art results.

However, training contextual representations is known to be a  resource-hungry process. For example, ELMo was reported to take about two weeks to train on a one-billion-token corpus with a vocabulary of 800,000 words using three GPUs\footnote{https://github.com/allenai/bilm-tf/issues/55}. This slow training procedure hinders the development cycle, prevents fine-grained parameter tuning, and makes training contextual representations inaccessible to a broader community. More importantly, the success of these models stems from the large amount of data they used. It is challenging, if not impossible, to train a contextual representation model on a larger corpus with tens or hundreds of billions of tokens.

In this work, we explore how to accelerate contextual representation learning. We target the softmax layer as the major cause of inefficiency. This component takes up a huge portion of all trainable parameters (80\% for ELMo) and consumes a huge amount of training time. However, this layer will be discarded in the final model as the goal of contextual representation learning is to build a generic encoder. Therefore, it is rather a waste to allocate extensive computational resources to the softmax layer to obtain the best prediction of masked words.
As mentioned, language modeling can be viewed as predicting the missing words based on a context vector generated by the contextual encoder and an embedding matrix $\bm{W}$ for target words \cite{tie_embedding, tie_embedding_2}.
Learning the contextual encoder is difficult while learning word representation has been extensively studied \cite{word2vec}. It is natural to use a pre-trained word embedding to replace $\bm{W}$ and thus decouple learning contexts and words. 

In this paper, we propose an efficient framework to learn the contextual encoder by leveraging pre-trained word embeddings\footnote{The code and models will be released in the near future.}. Instead of using a softmax layer to predict the distribution of the missing word, we utilize and extend the \semfit{} layer \cite{kumar2018mises} to predict the embedding of the missing word. In the training process, the contextual encoder is learned by minimizing the distance between its output and a pre-trained target word embedding. The \semfit\ layer, with constant time complexity and small memory footprint, perfectly serves our desire to decouple learning contexts and words and devote most computational resources to the contextual encoder.
Our contributions are as follows:
\begin{compactitem}
    \item We introduce the \semfit\ layer into contextual representation learning and further improve it with open-vocabulary word embeddings. The resulting model is computationally efficient and can be trained with an untruncated vocabulary. (Section \ref{sec:apprach})
    \item We discuss the global objective of the \semfit{} layer and draw a connection to dimension reduction. We show that the \semfit\ layer is particularly suitable for contextual representation learning. (Section \ref{sec:semfit})
    \item We empirically show that our approach significantly reduces the training time of ELMo, while maintaining competitive performance on most of the end tasks. (Section \ref{Experiment})
    \item We conduct a thorough analysis of our approach, discussing different modeling choices and computational efficiency. We also analyze the subword language model, a strong baseline used in GPT and BERT to circumvent the problem of large vocabulary. (Section \ref{Analysis})
\end{compactitem}

\section{Related Work}\label{sec:relatedwork}

\paragraph{Contextual Representation}
We review recently proposed contextual representation models from two aspects: how they are trained and how these pre-trained models are used in downstream tasks. 

CoVe \cite{cove} trained a machine translation model, and used the source language encoder as a contextual representation model for other downstream tasks. As large in-domain parallel corpus is hard to obtain, the potential of CoVe is limited. In contrast, a few recent approaches learn contextual encoder on unannotated corpus with language model objectives.  ELMo \cite{elmo} concatenated a forward and a backward LSTM-based language model while GPT-1 \cite{GPT} and GPT-2 \cite{GPT2} used unidirectional transformer-based language models. BERT \cite{bert} introduced masked language models and provided deep bidirectional representation.

There are mainly two existing strategies for applying pre-trained contextual representations to downstream tasks: 1) \textit{feature-based} and 2) \textit{fine-tuning}. For the \textit{feature-based} approach (e.g., ELMo, CoVe), fixed features are extracted from the contextual encoder and inserted into task-specific architectures.
In the \textit{fine-tuning} approach (e.g., BERT, GPT-1), the contextual encoder is designed as a part of the network architecture for downstream tasks, and its parameters are fine-tuned with downstream task data. BERT was designed for the \textit{fine-tuning} approach but it was also evaluated with the \textit{feature-based} approach. GPT-2 is a scaled-up version of GPT-1 and exhibits strong performance under zero-shot settings.

\paragraph{The Large Vocabulary Issue}
The large and ever-growing vocabulary has been considered an obstacle to scaling up language models. We review existing solutions to this issue from both the language modeling and contextual representation literature.

Most studies for language modeling focus on directly reducing the complexity of the softmax layer. Following \newcite{kumar2018mises}, we roughly group them into two categories: sampling-based approximations and structural approximations. Sampling-based approximations include the sampled softmax \cite{sampled_softmax} and NCE \cite{NCE}. The sampled softmax approximates the normalization term of softmax by sampling a subset of negative targets, while NCE replaces the softmax with a binary classifier. On the other hand, structural approximations such as the hierarchical softmax \cite{Hierachical_Softmax} and the adaptive softmax \cite{adaptive_softmax}, form a structural hierarchy to avoid expensive normalization. The adaptive softmax, in particular, group words in the vocabulary into either a short-list or clusters of rare words.
For frequent words, a softmax over the short-list would suffice, which reduces computation and memory usage significantly. The adaptive softmax has been shown to achieve results close to that of the full softmax whilst maintaining high GPU efficiency \cite{Merity2018LMScale}.

Regarding contextual representation models, ELMo used the sampled softmax while GPT and BERT resorted to subword methods. Specifically, they used WordPiece \cite{google_wordpiece} or BPE \cite{subword} to split the words into subwords and the language models were trained to consume and predict these subwords. This method is efficient and scalable, as the subword vocabulary can be kept small. Language models trained in this way are neither strictly \textit{word-level} nor \textit{character-level}. Hence we categorize them as \textit{subword-level} language models. One potential drawback, however, is that these models produce representations for fragments of words, and it takes extra efforts to generate word-level representations from them. 
In this paper, we focus on \textit{word-level} language models and we will discuss \textit{subword-level} language models in Section \ref{Exp:Subword}.

\section{Approach}\label{sec:apprach}

In this section, we illustrate our approach to accelerating the training process of contextual representation models, where the goal is to build a generic contextual encoder.

We first review the procedure for learning a contextual encoder. Given a sequence of words $\{w_1, w_2, \dots, w_n\}$, the encoder seeks to form a rich contextual representation for every word based on their surrounding words, i.e. the context. An input layer, which is usually a word embedding or a character-CNN \cite{kim2016cnn}, produces a context-insensitive word representation $\bm{x_i^0}$ for word $w_i$. Then, $\bm{x_i^0}$ goes through a K-layer contextualizing structure, such as LSTM \cite{hochreiter1997lstm}, Gated CNN \cite{GatedCNN}, or Transformer \cite{Transformer}. Each layer outputs a context-dependent vector $\bm{x_i^k} (k = 1, \dots ,K)$\footnote{For example, the context-dependent vectors are the hidden states if LSTM is used as the contextualizing structure.}. The top layer's output $\bm{x_i^K}$ is the final output of the encoder. Notice that when using this contextual encoder in downstream tasks, recent methods have gone beyond simply using the top layer output. They either combine different layers' outputs $\bm{x_i^k}$ to obtain hierarchical representation (ELMo) or fine-tune the whole encoder (GPT). To avoid confusion, we refer to the contextual information captured by the encoder as the \textbf{contextual representation}. We denote the top layer's output vectors $\bm{x_i^K}$ as $\bm{c_i}$ and refer to it as the \textbf{context vector}. 

Training the contextual encoder involves embedding a context $c$ into a context vector $\bm{c}$ and using it to predict a missing word\footnote{In unidirectional language models, for word $w_i$, $\bm{c_i}$ is only dependent on $w_{1\sim i}$ or $w_{i\sim n}$, and we train the $\bm{c_i}$ to predict $w_{i+1}$ or $w_{i-1}$. For the masked language model in BERT, part of the text is masked and the task is to predict the masked words. Suppose word $w_i$ is masked, $\bm{c_i}$ is actually dependent on the whole masked sentence and we train $\bm{c_i}$ to predict $w_i$.}. The conventional way is to attach a softmax layer to the contextual encoder. The softmax layer multiplies $\bm{c}$ with an embedding matrix $\bm{W} \in \mathcal{R}^{V \times m}$ and then uses a softmax function to produce a conditional distribution over the vocabulary. Suppose the target word is $w$ and the context is $c$, the learning objective $l(c,w)$ for each instance is maximizing the negative log-likelihood of $w$ under $c$:
\begin{eqnarray}
    l(c,w) & = &- \log p(w|c) \nonumber \\
      & = &- \log softmax(\bm{cW^{T}}) \nonumber \\
      & = & - \bm{c} \cdot \bm{w} + \log\sum\nolimits_{w'}{\exp{\bm{c} \cdot \bm{w'}}} \label{Eq:LMObj}
\end{eqnarray}
$\bm{w} \in \mathcal{R}^{m}$ is a row from $\bm{W}$ corresponding to $w$ and the second term sums over the vocabulary. $V$ is the size of the vocabulary while $m \ll V$ is the size of the context vector. The overall learning objective is 
\begin{equation}
    \mathcal{L} = \sum\nolimits_{(c,w)} \#(c,w) l(c,w)
\end{equation}
where $\#(c,w)$ is the number of occurrences of the pair $(c,w)$ in the  corpus.
Note that the size of $W$ and the computational complexity of the normalization term (the second term in Eq. \eqref{Eq:LMObj}) scale linearly to the vocabulary size. Therefore, when the vocabulary size is large, the vanilla softmax layer becomes the bottleneck. Many approaches (Section \ref{sec:relatedwork}) have been proposed to address this issue, but we think there is still room for improvement.

We accelerates the training by replacing the softmax layer with a \semfit\ layer. Instead of maximizing the log-likelihood, after embedding the context into $\bm{c}$, we directly minimize the distance between the context vector $\bm{c}$ and a \textbf{target word embedding} $\bm{w}$,
\begin{equation}
l(c,w) = d(\bm{c}, \bm{w}).
\end{equation}
Notice that the target word embedding $\bm{w}$ is pre-trained and fixed. The distance function $d$ could be the L2 distance, cosine distance or some probabilistic distance metrics (see discussions in Section \ref{sec:semfit}).
For the rest of this section, we will discuss the advantages of the proposed approach in contextual representation learning.

\subsection{Computational Efficiency}
\label{subsec:computational_efficiency}
We analyze the computational efficiency of the proposed approach against existing acceleration approaches for the softmax layer. In particular, we discuss the sampled softmax and the adaptive softmax, both being popular choices for speeding up softmax \cite{limitslm, adaptive_softmax, Merity2018LMScale}.

\paragraph{Computational Time Complexity} 
In the \semfit\ layer, we only need to calculate the distance between two $m$-dimensional  vectors. Without the normalization term or the need to sample words, the \semfit\ layer has $O(1)$ time complexity, which grants scalability with respect to the vocabulary size. The time spent on the \semfit\ layer remains constant and negligible regardless of the vocabulary size. 
In comparison, the time complexity of the sampled softmax is proportional to the number of negative samples per batch. When the vocabulary is huge, a large number of negative samples are needed \cite{limitslm}. For the adaptive softmax, the time complexity is determined by the capacities of the short-list and rare-word clusters, which grows sub-linearly with respect to the vocabulary size.

\paragraph{Trainable Parameter Size}
The softmax layer takes up a huge part of the parameters of a language model. Here, we list the parameter sizes of models reported in the literature trained on the One Billion Word Benchmark \cite{OneBillionWord}, which is also the corpus ELMo was trained on. For ELMo, the character-CNN and LSTM have about 100 million parameters while the softmax layer has 400 million parameters. For the bigLSTM in \newcite{limitslm}, the softmax takes up 840 million parameters, while all other parts have 182 million parameters. These models used the sampled softmax, which is only designed to accelerate calculating the normalization term and does not reduce the trainable parameter size. The adaptive softmax proposes to reduce $m$ for rare words. The trainable parameter size is effectively reduced but still remains sizable. For a model trained on the same corpus \cite{adaptive_softmax}, the LSTM has around 50 million parameters while the adaptive softmax still amounts to 240 million parameters.
Our approach, on the other hand, uses a pre-trained word embedding, reducing the trainable parameters of the output layer from hundreds of millions to almost zero.

\paragraph{GPU Memory Efficiency}
Our approach exhibits exceptional GPU memory efficiency, due to reductions of the computational time complexity (with fewer intermediate results to keep) and trainable parameter size (with fewer parameters to store\footnote{We keep the pre-trained embedding in the main memory instead of loading them to the GPU memory since it does not need to be updated. This comes at a cost as we need to move the embedding needed for words in a batch from CPU to GPU at each time step. When the GPU memory is abundant, we could keep the word embedding on GPU to avoid this additional communication cost. But on mainstream hardwares, the GPU memory is often comparatively limited and this trick proves to be beneficial in our preliminary experiments.}). As a result, compared to the adaptive softmax, models with our technique are 2 to 5 times more memory efficient (Section \ref{Exp:DetailedEfficiency}). The memory efficiency further adds up to the speed advantage, because loading more words in each batch allows better parallelism when utilizing GPUs.

\paragraph{Scalability Across GPUs}
To scale up modern deep learning systems, frameworks are designed to train models with synchronous SGD with very large batch size and consequently a large number of GPUs \cite{KaimingImagenetOneHour,ImagenetInMinuetes}.
However, the overhead of running on multiple GPUs is the communication cost spent on synchronizing the parameters and their gradients across machines, which is proportional to the size of parameters that need to be updated. For the sampled softmax, due to the use of the sparse gradient, the communication cost is proportional to the number of the sampled words. For the adaptive softmax, since full softmax is still used within the short-list and each cluster, the sparse gradient trick is not available and the communication cost is proportional to the trainable parameter size. The \semfit\ layer, on the other hand, incurs little communication cost, making it more efficient to train models on multiple GPUs.

\paragraph{Easily-accessible Word Embedding}
With all these computational advantages, the only prerequisite for the \semfit\ layer is a pre-trained word embedding. Fortunately, learning word embeddings is much cheaper than learning contextual representations. For example, training a FastText embedding \cite{fasttext} on the One Billion Word Benchmark took merely two hours on an average CPU machine. Moreover, several existing word embeddings trained on large corpora in different domains are publicly available.

\subsection{Open-vocabulary Word Embedding}
Theoretically, we can use any pre-trained word representation as the target word embedding for the \semfit\ layer. We exploit a particular kind of word representation, the open-vocabulary word embedding, such as the FastText embedding and the \textsc{mimick} model \cite{mimick}. These embeddings utilize character or subword information to provide embedding for out-of-vocabulary (OOV) words. Combining the \semfit\ layer with open-vocabulary embedding, we can train contextual encoders with untruncated vocabulary while making substantial simplifications to the input layer.

\paragraph{Scalable Word Representation}
In the vanilla softmax, a simple $V$-by-$m$ matrix $\bm{W}$ is used to provide word representation. However, when we further scale up current language models, the parameter size of this matrix could become intractable. The adaptive softmax proposes to reduce $m$ for rare words such that the parameter size grows only sub-linearly. Here, we take another route and propose to utilize the open-vocabulary word embedding, which could represent an unlimited number of words with a fixed number of parameters, providing the much-needed scalability.

\paragraph{Learning with An Untruncated Vocabulary}
Combining the $O(1)$ time complexity and scalable word representation, we can conduct training with an untruncated vocabulary. Softmax-based methods keep a vocabulary to calculate the normalization term. With the \semfit{} layer, we only need the target word embedding and the context vector to conduct training. There is no need for truncating the vocabulary or keeping one because we could use multi-process to dynamically prepare the embedding.

According to \newcite{limitslm}, the ability to model rare words is an essential advantage of the neural models against N-gram models. Now with an untruncated vocabulary, we possess the power to model unlimited rare words. This feature is especially attractive if we are training on domains or languages with a long tail, for example, the biomedical domain where truncating the vocabulary may not be acceptable.

\paragraph{Open-vocabulary Input Layer}
To be easily applied in various tasks, the contextual encoder usually has an open-vocabulary input layer. ELMo used a character-CNN but it is relatively slow.
Thus we propose to reuse the pre-trained open-vocabulary word embedding as the input layer of the contextual encoder, reducing the time complexity of the input layer to $O(1)$. This also aligns with the main spirit of our approach, which is to spend computational resources on the most important part, the contextualizing structure like LSTM.

\section{The \semfit\ Layer}
\label{sec:semfit}
Though the \semfit\ layer is intuitive, its properties are less known.
In this section, we provide an analysis on the \semfit\ layer, extending the work of \newcite{kumar2018mises}. 
We investigate the global objective of different distance functions and link the \semfit\ layer to probabilistic language modeling and dimension reduction, which further justifies the intuition behind the \semfit\ layer. Finally, we discuss why the \semfit\ layer is particularly suited for contextual representation learning.

\subsection{Global Objective}\label{subsec:GlobalObjective}
Recall that the \semfit\ layer minimizes the distance between the context vector $\bm{c}$ and the target word embedding $\bm{w}$, $ l = d(\bm{c}, \bm{w})$.
There are several choices regarding the distance function $d$. We investigate these losses with different distance functions from \newcite{kumar2018mises} \footnote{For L2 loss, the $\bm{{w}}$ is the unnormalized word embedding while for cosine and NLLvMF loss, the $\bm{{w}}$ is the normalized word embedding. For simplicity, we abuse the notation and use $\bm{{w}}$ uniformly. $\bm{\bar{c}}$ is the normalized $\bm{{c}}$.}:
\begin{compactitem}
    \item L2: $(\bm{{c}} - \bm{{w}})^2$
    \item Cosine: $-{\bm{\bar{c}} \cdot \bm{{w}}} $
    \item NLLvMF: $ -\log C_m(\|\bm{c}\|) -
\lambda_2 \bm{{c}} \cdot \bm{{w}} + \lambda_1 \|\bm{c}\| $
\end{compactitem}

Although the above losses have different interpretations, we find that they are similar. Specifically, we can rewrite the global objective in an uniform way\footnote{$\bm{\tilde{c}}$ is a vector corresponding to the context $c$. $\bm{\tilde{c}}$, $f(\|\bm{\tilde{c}}\|))$ and $\lambda$ take different forms in different losses.
In the L2 loss, $\bm{\tilde{c}}$ is the unormalized $\bm{{c}}$, $f(\|\bm{\tilde{c}}\|)) = \bm{{c}}^2$, and $\lambda = 2$. In the cosine loss, $\bm{\tilde{c}}$ is $\bm{\bar{c}}$, $f(\|\bm{\tilde{c}}\|)) = 0$, and $\lambda = 1$. In NLLvMF, $\bm{\tilde{c}}$ is the unormalized $\bm{{c}}$ while $\lambda = \lambda_2$.  $f(\|\bm{\tilde{c}}\|)) = -\log C_d(\|\bm{c}\|) + \lambda_1 \|\bm{c}\|$ serves as a way to constrain the norm of $\bm{{c}}$.}:
{\small
\begin{equation*} 
\begin{aligned}
& \mathcal{L}  = \sum_{(c, w)} \#(c, w) (-\lambda \bm{\tilde{c}} \cdot \bm{{w}} + f(\|\bm{\tilde{c}}\|)) \\
  & = -\lambda \sum_{c} \bm{\tilde{c}} \cdot\sum_{w}\#(c, w) \bm{{w}} +  \sum_{(c,w)}\#(c, w)f(\|\bm{\tilde{c}}\|) \\
  & = -\lambda \sum_{c} \#(c)\bm{\tilde{c}}\! \cdot\! \sum_{w} p(w|c) \bm{w} \!+\!  \sum_{(c,w)}\!\#(c, w)\!f(\|\bm{\tilde{c}}\|) \\
\end{aligned}
\end{equation*}
}

The first term minimizes the inner-product between the contextual encoder output and the target embedding, and the second term adds different constraints on the encoder output's norm. Although for different losses the norm of the optimal $\bm{c}$ might be different, the direction of the optimal $\bm{c}$ should be the same as $\Sigma_{w} p(w|c) \bm{w}$. Simply put, the \semfit\ layer models the weighted average word embedding under a context.

\subsection{Dimension Reduction}\label{subsec:DimensionReduction}
We seek to justify the very idea of the \semfit\ layer. We want to answer why minimizing the distance between the encoder output and the target word embedding should work.

We first show that the \semfit\ layer is essentially performing language modeling after dimension reduction. Following \newcite{levy2014neural} and \newcite{Yang2017SoftmaxBottleneck}, language modeling can be viewed as modeling a conditional probability matrix $\bm{P} \in \mathcal{R}^{N \times V}$, where $N$ is the number of all possible contexts and $V$ is the vocabulary size. Each row of $\bm{P}$ corresponds to the conditional distribution of the word under a certain context. Softmax-based methods seek to find a $\bm{C} \in \mathcal{R}^{N \times m}$ and $\bm{W} \in \mathcal{R}^{V \times m}$, such that $softmax(\bm{C}\bm{W^{T}})$ best approximates $\bm{P}$. For the \semfit\ Layer, we are modeling $\Sigma_{w} p(w|c) \bm{w}$, which translates to $\bm{P}\bm{W}$ in the matrix form. $\bm{W}$ is pre-trained and fixed. We are essentially conducting ``multivariate regression'' on $\bm{P}$ after dimension reduction with $\bm{W}$ as the projection matrix. 

So the question becomes, what matrix serves as a good projection matrix and why a pre-trained word embedding would be a good choice? When doing dimension reduction, we either strive to preserve the most variance (PCA) \cite{PCA} or achieve the least reconstruction error (SVD) \cite{SVD}.
Suppose we have access to $\bm{P}$, we could easily perform PCA or SVD. Concretely, one could perform SVD on $\bm{M} \in  \mathcal{R}^{N \times V}$, and get
$  \bm{M_m} = \bm{U_m} \bm{\Sigma_m} \bm{V_m^T},$
where $\bm{U_m} \in \mathcal{R}^{N \times m}$, $\bm{\Sigma_m} \in \mathcal{R}^{m \times m}$ and $\bm{V_m} \in \mathcal{R}^{V \times m}$. In PCA, $\bm{M}$ is centered $\bm{P}$ and in SVD, $\bm{M}$ is the raw $\bm{P}$. $\bm{M_m}$ is the matrix of rank $m$ that best approximates the original matrix $\bm{M}$. $\bm{V_m^T}$ is the optimal projection matrix $\bm{W}$ in either case. In practice, though we could not get the full $\bm{P}$, we could use a simplified definition of context and get an approximated $\bm{P}$.

Interestingly, using a simplified definition of context is a common practice in learning word embedding \cite{levy2014neural}. SVD is also a recognized method to construct word embedding. When $\bm{M}$ is the PMI matrix or the conditional probability matrix, $ (\bm{\Sigma_d})^\alpha \bm{V_d}$ serves as a good word embedding \cite{levy2014neural}, where $\alpha \in [0, 1]$ is a tunable parameter.

This finding shows that there exists a natural link between the \semfit\ layer and SVD word embedding. Moreover, we could intuitively see that other kinds of word embedding might also serve well as the projection matrix $\bm{W}$. Suppose we want to preserve the variance of $\bm{P}$ in the spirit of PCA. If two contexts have very different conditional distribution, then $\Sigma_{w} p(w|c) \bm{\tilde{w}}$ for these two contexts should have very different directions, assuming that dissimilar words have dissimilar word vectors. Thus, a pre-trained word embedding might help us preserve much of the variance. 

\subsection{Decoding Algorithm}\label{subsec:DecodingAlgorithm}
The above analysis strengthens our argument that the \semfit\ layer especially suits contextual representation learning. 
To illustrate, one could see that the \semfit\ layer may not excel at predicting the next word, which in theory is determined by its ability to approximate $\bm{P}$. The \semfit\ layer models  $\bm{PW}$, and unless we get the optimal $\bm{W}$,  it is hard to get an approximated $\bm{P}$ from $\bm{PW}$; therefore, we cannot calculate perplexity from the \semfit\ layer. This could cause problems to tasks like machine translation or text generation where the ability to predict the next word is essential. For contextual representation, there exists no such problem as we do not seek to induce $\bm{P}$. 

As a way to predict the next word with the \semfit\ layer, \newcite{kumar2018mises} proposed to search for the nearest neighbor of the context vector $\bm{c}$ in the target embedding space $\bm{W}$. This decoding algorithm is sub-optimal unless the distribution of $p(w|c)$ is sharp. Suppose our training resulted in the global optimum, finding the nearest neighbor of $\bm{c}$ is equivalent to finding the word vector with the largest inner product with $\bm{p(w|c)}\bm{W}$\footnote{This is true for the cosine and NLLvMF distance. For L2 distance it is more complicated and we mainly focus on the cosine and NLLvMF distance here. But note that the decoding algorithm is still flawed under L2 distance.}.
For a specific word $w_0$, its inner product
\begin{eqnarray}
   d_{w_0} & = & \bm{c_{optim}} \cdot \bm{w_0} \nonumber \\
   & \propto & \bm{p(w|c)}\bm{W} \cdot \bm{w_0} \nonumber \\
   & = & p(w_0|c) + \Sigma_{w'\neq w_0} p(w'|c) \bm{w}' \cdot \bm{w_0} \quad \label{equation:dot}
\end{eqnarray}
$d_{w_0}$ is influenced not only by $w_0$'s conditional probability, but also by its distance to other words and their conditional probabilities. As a result, the most desired word does not always get chosen. Rewriting this in a matrix form, we are essentially calculating $\bm{p(w|c)}\bm{WW^T}$ during decoding. Unless $\bm{W}$ is from the SVD of $\bm{P}$, the resulting matrix is not an approximation for $\bm{P}$.
Note that in some applications, such as machine translation, the conditional distribution $p(w|c)$ is often sharp. In this case, the first term in Eq. \eqref{equation:dot} dominates over the second term and the approximation is valid.

\newcommand{\original}{\textsc{ELMo$_{\small \textsc{Org}}$}}
\newcommand{\adaptive}{\textsc{ELMo\textsl{-A}$_{\small \textsc{}}$}}
\newcommand{\ourcc}{\textsc{ELMo}\textsl{-S}$_{\small \textsc{}}$}
\newcommand{\ourone}{\textsc{ELMo\textsl{-S}$_{\small \textsc{OneB}}$}}
\newcommand{\ourcnn}{\textsc{ELMo\textsl{-S}$_{\small \textsc{CNN}}$}}
\newcommand{\base}{\textsc{Base}}
\newcommand{\rerun} {\textsc{ELMo}}
\newcommand{\fasttextcc}{\textsc{FastText$_{\small \textsc{cc}}$}}
\newcommand{\subword}{\textsc{ELMo\textsl{-Sub}}}

\section{Experiment}\label{Experiment}
The proposed approach is generic and can be applied to accelerating the training process of word-level contextual representation models. In this section, we take ELMo as an example and demonstrate that our approach significantly speeds up ELMo, while largely maintaining its performance.

\subsection{Setup}
ELMo consists of a forward and a backward language model, trained on the One Billion Word Benchmark for 10 epochs.
For a fair comparison, all models we introduce are trained on the same corpus for 10 epochs. All experiments are conducted on a workstation with 8 GeForce GTX 1080Ti GPUs, 40 Intel Xeon E5 CPUs, and 128G main memory. The training code is written in PyTorch such that we could evaluate on most downstream tasks with AllenNLP \cite{AllenNLP}.

\paragraph{Models}
\begin{table*}
\centering
\small
\begin{tabular}{l|lll}
\toprule
Model & 
Input & 
Contextualizing & 
Output\\ 
\midrule
\textsc{{\rerun{}}}  &   CNN & LSTM  & Sampled Softmax   \\
\textsc{{\ourcc{}}} &  \fasttextcc{} & LSTM w/ LN  & \semfit{} w/ \fasttextcc{} \\
\textsc{{\adaptive{}}}   & \fasttextcc{} & LSTM w/ LN  & Adaptive Softmax\\
\midrule 
\textsc{{\ourone{}}}   & {\textsc{FastText$_{\small \textsc{OneB}}$}}  & LSTM w/ LN & \semfit{} w/ {\textsc{FastText$_{\small \textsc{One}}$}}\\
\textsc{{\ourcnn{}}}   & Trained CNN  & LSTM w/ LN  & \semfit{} w/ Trained CNN \\
\textsc{{\subword{}}}   & Subword & LSTM w/ LN & Softmax\\
\bottomrule
\end{tabular}
\caption{Specifications of ELMo models we introduce in Section \ref{Experiment} and Section \ref{Analysis}.}
\label{table:models}
\end{table*}
To verify the efficacy of the proposed method, we introduce an ELMo model trained with our acceleration approach (\ourcc{}). We also include the original ELMo (\rerun{}) for comparison. Our main proposal is to use the \semfit\ layer as the output layer but there are substantial differences between \rerun{} and \ourcc{} besides the output layer. To make a fair comparison, we introduce an ELMo model with the adaptive softmax (\adaptive{}). \adaptive{} is designed to differ from \ourcc{} only in the output layer so that we could study the effect of the \semfit\ layer in isolation.

In the following, we describe the details of these models mainly from three aspects: 1) the input layer, 2) the contextualizing structure and 3) the output layer. A brief summary of these models are available in Table \ref{table:models}.
\begin{compactitem}
    \item \textbf{\rerun{}}: The input layer is a character-CNN, and the contextualizing structure is an LSTM with projection \cite{SakSB14LSTMProjection}. The output layer is a sampled softmax with 8192 negative samples per batch. This model is provided in AllenNLP by \newcite{elmo}.
    
    \item  \textbf{\ourcc{}}: The input layer is the FastText embedding trained on Common Crawl \cite{fasttext_resource}, denoted as \fasttextcc{}\footnote{There are two ways to use the FastText embedding. For words in a pre-defined vocabulary, one could use the conventional word embedding; for OOV words, one could use subword embedding to compute word embedding. For consistency, we always use the subword embedding.}. The contextualizing structure is an LSTM with projection the same size as the one in \rerun, but we added layer norm \cite{LayerNormalization} after the projection layer as we find it beneficial. The output layer is the \semfit{} layer with \fasttextcc{} embedding as the target embedding. We use the cosine distance as the distance function of the \semfit{} layer because it is free of hyper-parameters and we find that it gives a satisfying and stable performance in preliminary experiments. The learning rate schedule from \newcite{BestOfBothWorlds} is used to aid large-batch training.
    
    \item \textbf{\adaptive{}}: The input layer, the contextualizing structure and the training recipe of \adaptive{} is kept the same as \ourcc{}. The only difference is that the output layer of \adaptive{} is an adaptive softmax with 120 million parameters, half of the size of the one reported in \newcite{adaptive_softmax} on the same corpus. \adaptive{} achieves a perplexity of 35.8, 3.9 points lower than \rerun{}'s 39.7. This shows that it is an efficient yet strong baseline.

\end{compactitem}

We also list the performance of the following models for reference. \original\ and \base\ are results listed in \newcite{elmo} but they are tested using different configurations. \fasttextcc\ is the non-contextual word embedding trained on the Common Crawl corpus with 600 billion tokens, which serves as a baseline non-contextual model.

\begin{table}
\centering
\small
\begin{tabular}{lllll}
\toprule
Model & 
Time &
SpeedUp &
Batch &
Params\\ 
\midrule

\rerun{} &   14 x 3 & 1x  & 128 & 499m \\
\adaptive{} &  5.7 x 4 & 1.8x  & 256 & 196m\\
\ourcc{} &  2.5 x 4 & 4.2x  & 768 &  76m \\

\bottomrule
\end{tabular}
\caption{Model Efficiency. Time is in Days x Cards format. Batch is the maximal batch size per card. Params is the total trainable parameters in millions.}
\label{table:elmo_model_time}
\end{table}

\begin{table*}[thb]
\centering
\scriptsize
\begin{tabular}{l@{\hskip8pt}l@{\hskip8pt}l@{\hskip8pt}l@{\hskip8pt}l@{\hskip8pt}l@{\hskip8pt}l@{\hskip8pt}l@{\hskip8pt}l@{\hskip8pt}l@{\hskip8pt}l@{\hskip8pt}l}
\toprule
{Task} & 
&
\original & 
\base & 
\fasttextcc &
& 
\rerun &
\adaptive & 
\ourcc &
&
\ourone &
\ourcnn \\

\cmidrule{1-1} \cmidrule{3-5} \cmidrule{7-9} \cmidrule{11-12}

SNLI  & &   88.7 & 88.0 & 87.7 & & 88.5    & 88.9     & 88.8 & & 88.4 & 88.2 \\
Coref & & NA & NA & 68.90  & & 72.9  & 72.9  & 72.9 & & 73.0 & 72.9\\
SST-5 & & 54.7 & 51.4 &  51.30 $\pm$ 0.77 &  & 52.96 $\pm$ 2.26   & 53.58 $\pm$ 0.77   & 53.80 $\pm$ 0.73 &   & 53.86 $\pm$ 4.02 & 53.38 $\pm$ 0.68 \\

\cmidrule{1-1} \cmidrule{3-5} \cmidrule{7-9} \cmidrule{11-12}

NER   & &   92.22  & 90.15 & 90.97 $\pm$ 0.43 &   & 92.51 $\pm$ 0.28   & 92.28 $\pm$ 0.20  & 92.24 $\pm$ 0.10 &   & 92.03 $\pm$ 0.47 & 92.24 $\pm$ 0.36 \\

SRL & &  84.6 & 81.4 & 80.2 &  & 83.4 & 82.7  & 82.4 &  & 82.2 & 82.8 \\

\bottomrule
\end{tabular}
\caption{Performance of main competing models and two ablation models on five NLP benchmarks. Due to the small test sizes for NER and SST-5, we report mean and standard deviation across three runs.}
\label{table:overall}
\end{table*}

\paragraph{Downstream Tasks}
We follow ELMo and use the \textit{feature-based} approach to evaluate contextual representations on downstream benchmarks.
ELMo was evaluated on six benchmarks and we conduct evaluations on five of them. SQuAD \cite{Rajpurkar2016Squad} is not available for implementation reasons\footnote{The SQuAD experiment in \newcite{elmo} was conducted with code written in TensorFlow. The experiment setting is not currently available in AllenNLP (https://github.com/allenai/allennlp/pull/1626), nor can it be easily replicated in PyTorch.}. We briefly review the benchmarks and task-specific models. For detailed descriptions, please refer to \newcite{elmo}.

\begin{compactitem}
    \item \textbf{SNLI}: The textual entailment task seeks to determine whether a ``hypothesis'' can be entailed from a ``premise''. The dataset is the SNLI dataset \cite{SNLIDataset} and the model is ESIM \cite{ESIM}.
    
    \item \textbf{Coref}: Coreference resolution is the task of clustering mentions in text that refer to the same underlying entities. The dataset is from CoNLL 2012 shared task \cite{CorefDataset} and the model is from \newcite{CorefNewModel}.
    
    \item \textbf{SST-5}: SST-5 \cite{SSTDataset} involves selecting one of five labels to describe a sentence from a movie review. The model is the BCN model from \newcite{cove}.
    
    \item \textbf{NER}: The CoNLL 2003 NER task \cite{NERDataset} consists of newswire from the Reuters RCV1 corpus tagged with four different entity types. The model is a biLSTM-CRF from \newcite{elmo}, similar to \newcite{Collobert2011Scratch}.
    
    \item \textbf{SRL}: Semantic role labeling (SRL) models the predicate-argument structure of a sentence, and is often described as answering ``Who did what to whom''. The model is from \newcite{SRLModel} and the dataset is from \newcite{SRLDataset}.
\end{compactitem}

For SNLI, SST-5, NER, and SRL, we used the same downstream models as in \newcite{elmo} re-implemented in AllenNLP. For Coref, \newcite{elmo} used the model from \newcite{CorefOldModel}. The authors from \newcite{CorefOldModel} later updated their model and achieved superior scores with the new model \cite{CorefNewModel}. Thus we adopted the improved model in our experiments on Coref.
For all the tasks, we use the default configuration with modest tuning, but all models are tested under the same configurations. Notice that the hyper-parameters are tuned to maximize the performance for the original ELMo and they may not be optimal for other models\footnote{For example, the number of epochs is tuned for ELMo and some models may need more epochs to train. In addition, for SRL, the reported score by AllenNLP is lower than the score from CoNLL official script.}.
But since all models are tested under the same hyper-parameters and our setting favorites the baseline ELMo model, the results still reflect the performance of our approach.

\subsection{Model Efficiency}\label{Exp:WholeEfficiency}
In Table \ref{table:elmo_model_time}, we report the computational efficiency of the models\footnote{The statistics of \rerun{} are reported by the authors of ELMo so the number of the GPU used is different for \rerun{}. We tested \adaptive{} and \ourcc{} using the same kind of GPU so the numbers are directly comparable. This setting actually favors \rerun{} as the communication cost on three cards is smaller than that on four cards.}. Overall, our simplifications to the input layer and the output layer of ELMo brings significant computational efficiency. \ourcc\ is 4.2x faster and 6x more memory efficient than \rerun{}.

To give a clear view of the speedup the \semfit\ layer brings, we compare \ourcc\ with \adaptive{}. \adaptive{} differs from \ourcc{} only in the output layer by using an adaptive softmax half the size of the one reported in \newcite{adaptive_softmax}. Still, \ourcc{} has a 2.3x speed advantage and is 3 times more memory efficient.

Moreover, the contextualizing structure used in these three models is an LSTM with projection written without cuDNN acceleration\footnote{There is no readily-available implementation of LSTM with projection in PyTorch with cuDNN acceleration now.}, which is slower than a cuDNN LSTM \cite{cuDNN} and much slower than other fast structures like QRNN \cite{QRNN}, SRU \cite{SRU}, Gated CNN or Transformer. \newcite{DissectingELMo} showed that such faster structures could deliver close performance and 3-5x speedup. Thus our approach might achieve even better speedup with faster structures.

\subsection{Performance on Downstream Tasks} 
Table \ref{table:overall} reports the downstream task performance of each representation model. We focus on the three main competing models (\rerun, \adaptive, \ourcc) in the middle columns.

Our approach (\ourcc{}) works especially well on semantic-centric tasks, such as SNLI, Coref, and SST-5. It shows competitive or even better performance than \rerun\ and \adaptive{}. However, for tasks that required a certain level of syntactic information, such as NER and SRL~\cite{He2018SRLSyntax}, \ourcc\ suffers from slight performance degradation, but it still holds a large advantage over the pure word embedding model, \fasttextcc{}.
We suspect that the performance degradation is due to the pre-trained embedding we used. Therefore, we conduct further analyses and discuss the results in Section \ref{Exp:SemanticSyntactic}.

\section{Analysis}
\label{Analysis}
We conduct further analyses regarding certain modeling choices of our approach, the subword-level language models and conduct a more detailed analysis of computational efficiency.

\subsection{Modeling Choices}

\paragraph{Sensitivity to the Pre-trained Embedding}\label{Exp:CorpusSize}
Our previous experiments are based on the word embedding trained on Common Crawl. In this experiment, we analyze how sensitive our approach is to the pre-trained word embeddings. We trained a FastText embedding on the One Billion Word Benchmark and denote an \ourcc{} model trained with this embedding as the input and target embedding as \ourone{}. Notice that the code we used to train the one-billion-word embedding lacks a few new features compared to the common-crawl embedding provided in \newcite{fasttext_resource}. The one-billion-word embedding and consequently \ourone\ might be further improved with the new features. Comparing it to \ourcc\ (Table \ref{table:overall}), we find that this model holds up surprisingly well, with only minor performance decrease. \ourone\ is competitive with \rerun{} on SNLI, Coref, and SST-5 while being inferior on NER and SRL, still better than \fasttextcc{}.

We especially note that this \ourone\ model does not enjoy any additional resources since the pre-trained embedding is trained on the same corpus ELMo was trained on and training only took two hours. The performance of this model, especially on SNLI, Coref, and SST-5, is attractive given that we have made substantial simplifications to the model.

\paragraph{Semantic Versus Syntactic}
\label{Exp:SemanticSyntactic}

In Section \ref{Experiment}, we observed that models with FastText embedding uniformly performed worse than ELMo on SRL, which relied heavily on syntactic information. We suspect that the FastText embedding might be weaker on capturing syntactic information, while \newcite{DissectingELMo} revealed that the CNN layer in ELMo is strong on capturing syntactic information. This motivates us to explore whether we could use syntactic-rich embedding and get better results on SRL.
We find that the trained CNN layer from ELMo, surprisingly, serves as a kind of syntactic-rich word embedding. When we use that CNN layer as a pre-trained word embedding to train a \ourcc{} model (denoted as \ourcnn{}), we observed a notable performance increase on SRL (Table \ref{table:overall}).

\ourcnn\ is also a natural extension to an intriguing idea, CNN softmax from \newcite{limitslm}. They proposed to use a CNN to provide word representation for the softmax layer. The attraction lies in that CNN is very parameter-efficient and also posses the open-vocabulary feature. 
However, \newcite{limitslm} pointed out that the CNN softmax sometimes cannot differentiate between words with similar spellings but different meanings, which could potentially explain why \ourcnn{} is inferior on certain semantic tasks (SNLI and SST5).

\subsection{Subword-level Language Models}\label{Exp:Subword}
\begin{table*}[thb]
\centering
\scriptsize
\begin{tabular}{lllllllllllllll}
\toprule

\multirow{2}{*}{Models} 
& \multirow{2}{*}{SNLI} & \multirow{2}{*}{Coref} &  \multirow{2}{*}{SST-5} & \multirow{2}{*}{NER} & &
\multicolumn{6}{c}{SentEval} \\
& & & & & & \textbf{Avg} & SST-5 & SST-2 & TREC & MR & SUBJ 
\\
\midrule
\subword\ & 87.1 & 72.4 & 53.02 $\pm$ 2.08  & 92.17 $\pm$ 0.56
        & & 80.30 & 45.25 & 84.42 & 92.13 & 78.40 & 93.83   \\

\ourone\ & 88.4 & 73.0 & 53.86 $\pm$ 4.02 & 92.03 $\pm$ 0.47
        & & 80.31 &44.99 & 83.21 & 91.60 & 78.85 & 93.84 \\

\ourcc\ & 88.8 & 72.9 & 53.80 $\pm$ 0.73 & 92.24 $\pm$ 0.10
        & & 80.96 & 45.22 & 85.45 & 92.87 & 80.05 & 94.11 \\

\fasttextcc\ & 87.7 & 68.90 & 51.30 $\pm$ 0.77  &  90.97 $\pm$ 0.43
        & & 71.10 & 38.40 & 80.12 & 71.80 & 74.30 & 89.06\\

\bottomrule
\end{tabular}
\caption{Performance of models including \subword\ on downstream tasks. We show the average score of ten classification tasks from SentEval and showcase five of them.}
\label{table:subword}
\end{table*}

\begin{table*}[thb]   
\centering
\small
\begin{tabular}{lcccccc}
\toprule
 & Vocab & 
TimeBatch16  & 
TimeOneCard & 
TimeFourCard & 
Params & 
Batch
\\ 

\midrule 

\semfit\ & $\infty$ & 0.03s & 13.04s & 4.61s & 20m & 5120 
\\
\midrule
\multirow{3}{*}{\textsc{Adaptive}}
& 40K   & 0.05s & 24.78s & 9.68s  & 106m  & 2048\\
& 800K  & 0.06s & 29.99s & 16.14s  & 265m & 1024\\
& 2000K & 0.10s & 54.77s & 92.66s & 420m & 160 \\

\bottomrule
\end{tabular}
\caption{Statistics on the computation efficiency of the \semfit\ layer and the adaptive softmax. Time is in second. Params: Number of trainable parameters of the whole model in millions. Batch: Maximal batch size per card.}
\label{table:speed}
\end{table*} 

Next, we discuss the advantages and potential disadvantages of subword-level language models. By splitting the original words into smaller fragments (subwords), these models have a small vocabulary and can deal with arbitrary words, essentially circumventing the large vocabulary problem. They also possess the open-vocabulary feature in the input layer as they can split unseen words into seen subwords.

However, these models produce contextual representations for subwords rather than words. More concretely, consider a word ABC consisting of three subwords A, B, and C. Under the subword method, we would get a contextual representation for A, B, and C respectively. In some scenarios, we just want exactly one representation vector for word ABC instead of three. BERT approached this by using the representation for A as the representation for the whole word ABC. This method seems like an ad-hoc workaround rather than a principled solution. We are concerned that this could be unsuitable in scenarios where precise word-level representation is required. For example, it might not be optimal to use the subword-level contextual representation with the \textit{feature-based} method in word-level tasks such as Coref.

We conduct a controlled experiment to verify our concern. We introduce a subword-level ELMo using the BPE segmentation with 30,000 merges, denoted as \subword{}. \subword{} differs from \ourcc{} in the input layer and the output layer. The input layer is a vanilla lookup-table-style subword embedding while the output layer is a full softmax layer. The input and softmax embedding are tied but trained from scratch. All other settings are kept the same as those of \ourcc{}. To make sure that \subword{} is a fair and properly-implemented baseline, we additionally introduce some sentence-level tasks with simple architectures that we think \subword{} should be good at. Concretely, we include classification tasks in SentEval \cite{SentEval}, which attaches a simple softmax classifier on top of sentence embeddings to train models for sentence-level tasks. We follow \newcite{ELMoSentenceEmbedding} to get sentence embeddings from contextual representations.

We find that \subword\ has inconsistent performance (Table \ref{table:subword}). On SentEval, SST-5, and NER, it has similar performance with \ourone{}. However, it lags behind on Coref and has almost breaking performance on SNLI. It even failed to outperform \fasttextcc\ on SNLI, a non-contextual model, which we considered baseline. These results are consistent with the observation in a recent work \cite{Kitaev2018parsingBERT}. They find that special design has to be made to apply BERT to constituency parsing because of the subword segmentation.

However, we notice that the scope of this experiment is limited. It is very likely that when the model is scaled up or used with the \textit{fine-tuning} method, the aforementioned issue is alleviated, as evidenced by the performance of GPT and BERT. It is hard to judge whether the subword-level language models will work well under a specific setting, and we leave that to future work.

It is noteworthy that our approach still holds a speed advantage over subword-level language models. In those models, a softmax layer is still needed to normalize over a relatively small but non-negligible vocabulary. Moreover, words are split into subwords, which increases the length of a sentence. In our experiment, training \subword\ takes 3.9 days on four cards, which is 1.56x slower than training \ourcc{}.

\subsection{Computational Efficiency}\label{Exp:DetailedEfficiency}
In this section, we aim to provide a detailed study on the computational efficiency of the \semfit\ layer. We follow the setting in \newcite{adaptive_softmax} to 
compare with the adaptive softmax. We use a uni-directional LSTM with 2048 units. The input embedding is fixed as in our previous experiments. We vary the vocabulary size and show the results in Table \ref{table:speed}.

We first explain some of the statistics we report:
\begin{compactitem}
\item TimeBatch16: Time needed to finish a single batch with 16 examples. This reflects the computational time complexity of models.
\item TimeOneCard: Time needed to process one million words on one GPU when using the maximal batch size. This reflects the actual running time of models on one GPU card and is affected by both the computational time complexity and GPU memory efficiency.
\item TimeFourCard: Time needed to process one million words on a machine with four GPUs when using the maximal batch size. This reflects the actual running time of models on four GPU cards and is included to study the communication cost across GPUs.
\end{compactitem}

We have the following observations:
\paragraph{Speedup and Batch Size} When the vocabulary size is small (e.g., 40K), the speed gain from replacing softmax with \semfit\ for each batch is small (reflected by TimeBatch16). However, \semfit{} still benefits from its memory efficiency. Therefore, by using a larger batch size, the overall speedup (TimeOneCard) is large. On the other hand, when the vocabulary size is larger, {\sc Adaptive} becomes slower, while the complexity of \semfit\ remains constant.   

\paragraph{Multi-GPU Scalability} The speed superiority of the \semfit\ layer is magnified when we move to multiple GPUs. The speedup of the \semfit\ layer on TimeFourCard is consistently higher than that on TimeOneCard. This will be very useful when we scale to dozens or hundreds of machines.

\paragraph{Super Large Vocabulary} The \semfit\ layer has a great advantage when the vocabulary is super large. In the 2,000K vocabulary test, the speedup on one card is already great. Using four cards for the adaptive softmax is even counter-productive as the communication cost exceeds the benefit of more GPUs. In this experiment, the hyper-parameters of {\sc Adaptive} are the same as those in the 800K vocabulary test. We note that this could be suboptimal for {\sc Adaptive} as one could choose to trade off accuracy for efficiency and change the hyper-parameters to allocate even less computational resources to the rare words. Notice that the 2,000K vocabulary is not an impractical setting. It is created on a tokenized 250-billion-word Common Crawl corpus \cite{CommonCrawl250Corpus}, which only covers words that appear more than 397 times.

\section{Conclusion and Future Work}
We introduced an efficient framework to learn contextual representation without the softmax layer. The experiments with ELMo showed that we significantly accelerate the training of the current models while maintaining competitive performance on various downstream tasks. We also provided a theoretical explanation on the \semfit\ layer. For future work, we are interested in extending our approach to other contextual representation models such as BERT and scale these models to larger datasets.

\section*{Acknowledgements}
We would like to thank Muhao Chen for helpful comments. We also thank Yulia Tsvetkov and Sachin Kumar for help with implementing the \semfit{} layer as well as Jieyu Zhao, Kenton Lee, and Nelson Liu for providing reproducible code for experiments.

\bibliography{tacl2018}
\bibliographystyle{acl_natbib}

\end{document}